\documentclass[conference]{IEEEtran}
\IEEEoverridecommandlockouts
\usepackage{cite}
\usepackage{amsmath,amssymb,amsfonts}
\usepackage{algorithmic}
\usepackage{graphicx}
\usepackage{graphics}
\usepackage{textcomp}
\newcommand{\p}{+}
\newcommand{\m}{-}
\usepackage{xcolor}
\def\BibTeX{{\rm B\kern-.05em{\sc i\kern-.025em b}\kern-.08em
    T\kern-.1667em\lower.7ex\hbox{E}\kern-.125emX}}
\usepackage{scalerel}
\usepackage{tikz}
\usepackage{tikz-3dplot}
\usepackage[utf8]{inputenc}
\usepackage{pgfplots}
\pgfplotsset{compat=newest}
\usepgfplotslibrary{groupplots}
\usepgfplotslibrary{dateplot}
\usetikzlibrary{svg.path}
\definecolor{orcidlogocol}{HTML}{A6CE39}
\tikzset{
orcidlogo/.pic={
\fill[orcidlogocol] svg{M256,128c0,70.7-57.3,128-128,128C57.3,256,0,198.7,0,128C0,57.3,57.3,0,128,0C198.7,0,256,57.3,256,128z};
\fill[white] svg{M86.3,186.2H70.9V79.1h15.4v48.4V186.2z}
svg{M108.9,79.1h41.6c39.6,0,57,28.3,57,53.6c0,27.5-21.5,53.6-56.8,53.6h-41.8V79.1z M124.3,172.4h24.5c34.9,0,42.9-26.5,42.9-39.7c0-21.5-13.7-39.7-43.7-39.7h-23.7V172.4z}
svg{M88.7,56.8c0,5.5-4.5,10.1-10.1,10.1c-5.6,0-10.1-4.6-10.1-10.1c0-5.6,4.5-10.1,10.1-10.1C84.2,46.7,88.7,51.3,88.7,56.8z};
}
}
\newcommand\orcidicon[1]{\href{https://orcid.org/#1}{\mbox{\scalerel*{
\begin{tikzpicture}[yscale=-1,transform shape]
\pic{orcidlogo};
\end{tikzpicture}
}{|}}}}

\usepackage[ruled,vlined]{algorithm2e}
\allowdisplaybreaks
\usepackage[official]{eurosym}
\usepackage{nomencl}
\makenomenclature
\renewcommand\nomgroup[1]{%
  \item[\bfseries
  \ifstrequal{#1}{I}{Indices}{%
      \ifstrequal{#1}{V}{Variables (lower-case letters)}{%
          \ifstrequal{#1}{C}{Constants}{%
              \ifstrequal{#1}{P}{Parameters (upper-case letters)}{%
                \ifstrequal{#1}{J}{Sets}{}%
              }%
          }%
      }%
  }%
]}
\newcommand{\labE}{(\!E\!)}
\newcommand{\labEC}{(\!\p\!)}
\newcommand{\labED}{(\!\m\!)}
\newcommand{\labD}{(\!D\!)}
\newcommand{\labF}{(\!F\!)}
\newcommand{\labV}{(\!V\!)}
\newcommand{\labW}{(\!W\!)}
\newcommand{\conNA}{N\!A}
\nomenclature[Cna]{$\conNA$}{Number of action indices;}
\newcommand{\conNB}{N\!B}
\nomenclature[Cnb]{$\conNB$}{Number of bus indices;}
\newcommand{\conND}{N\!D}
\nomenclature[Cnd]{$\conND$}{Number of demand indices;}
\newcommand{\conNE}{N\!E}
\nomenclature[Cne]{$\conNE$}{Number of ESS indices;}
\newcommand{\conNF}{N\!F}
\nomenclature[Cnf]{$\conNF$}{Number of feeder indices;}
\newcommand{\conNO}{N\!O}
\nomenclature[Cno]{$\conNO$}{Number of observation indices;}
\newcommand{\conNT}{N\!T}
\nomenclature[Cnt]{$\conNT$}{Number of time indices;}
\newcommand{\conNS}{N\!S}
\nomenclature[Cns]{$\conNS$}{Number of scenarios;}
\newcommand{\conNV}{N\!V}
\nomenclature[Cnv]{$\conNV$}{Number of PV units;}
\newcommand{\conNW}{N\!W}
\nomenclature[Cnw]{$\conNW$}{Number of WT units;}
\newcommand{\setA}{\mathbb{A}}
\nomenclature[Ja]{$\setA$}{Set of action indices $\{1,2,\dots,\conNA\}$;}
\newcommand{\setB}{\mathbb{B}}
\nomenclature[Jb]{$\setB$}{Set of bus indices $\{1,2,\dots,\conNB\}$;}
\newcommand{\setD}{\mathbb{D}}
\nomenclature[Jd]{$\setD$}{Set of demand indices $\{1,2,\dots,\conND\}$;}
\newcommand{\setE}{\mathbb{E}}
\nomenclature[Je]{$\setE$}{Set of ESS indices $\{1,2,\dots,\conNE\}$;}
\newcommand{\setF}{\mathbb{F}}
\nomenclature[Jf]{$\setF$}{Set of feeder indices $\{1,2,\dots,\conNF\}$;}
\newcommand{\setO}{\mathbb{O}}
\nomenclature[Jo]{$\setO$}{Set of observation indices $\{1,2,\dots,\conNO\}$;}
\newcommand{\setS}{\mathbb{S}}
\nomenclature[Js]{$\setS$}{Set of scenario indices $\{1,2,\dots,\conNS\}$;}
\newcommand{\setT}{\mathbb{T}}
\nomenclature[Jt]{$\setT$}{Set of time indices $\{1,2,\dots,\conNT\}$;}
\newcommand{\setV}{\mathbb{V}}
\nomenclature[Jv]{$\setV$}{Set of PV unit indices $\{1,2,\dots,\conNV\}$;}
\newcommand{\setW}{\mathbb{W}}
\nomenclature[Jw]{$\setW$}{Set of WT unit indices $\{1,2,\dots,\conNW\}$;}
\newcommand{\setX}{\mathbb{X}}
\nomenclature[Jx]{$\setX$}{Set of variables;}
\newcommand{\inda}{a}
\nomenclature[Ia]{$\inda\in\setA$}{Action index;}
\newcommand{\indb}{b}
\nomenclature[Ib]{$\indb\in\setB$}{Bus index;}
\newcommand{\indd}{d}
\nomenclature[Id]{$\indd\in\setD$}{Demand index;}
\newcommand{\inde}{e}
\nomenclature[Id]{$\inde\in\setE$}{ESS index;}
\newcommand{\indf}{f}
\nomenclature[If]{$\indf\in\setF$}{Feeder index;}
\newcommand{\indo}{o}
\nomenclature[Io]{$\indo\in\setO$}{Observation index;}
\newcommand{\inds}{s}
\nomenclature[Is]{$\inds\in\setS$}{Scenario index;}
\newcommand{\indt}{t}
\nomenclature[It]{$\indt\in\setT$}{Time index;}
\newcommand{\indv}{v}
\nomenclature[Iv]{$\indv\in\setV$}{PV unit index;}
\newcommand{\indw}{w}
\nomenclature[Iw]{$\indw\in\setW$}{WT unit index;}
\newcommand{\vareet}{{e}_{\inde\inds\indt}}
\newcommand{\vareetminusone}{{e}_{\inde(\indt\m1)\inds}}
\nomenclature[Ve]{$\vareet$}{Stored energy BSS $\inde$ (MWh);}
\newcommand{\varpec}{{p}^{\labEC}_{\inde\inds\indt}}
\newcommand{\varped}{{p}^{\labED}_{\inde\inds\indt}}
\nomenclature[Vpc]{$\varpec$/$\varped$}{Charging/Discharging active power in BSS $\inde$ (MW);}
\newcommand{\varpf}{{p}^{\labF}_{\indf\inds\indt}}
\nomenclature[Vpf]{$\varpf$}{Active power of feeder $\indf$ (MW);}
\newcommand{\varuec}{{u}^{\labEC}_{\inde\inds\indt}}
\newcommand{\varuecp}{{u'}^{\labEC}_{\inde\inds\indt}}
\newcommand{\varued}{{u}^{\labED}_{\inde\inds\indt}}
\newcommand{\varuedp}{{u'}^{\labED}_{\inde\inds\indt}}
\nomenclature[Vuec]{$\varuec$/$\varued$}{Charging/Discharging option binary variables $\varuec,\varued\in\{0,1\}$;}
\newcommand{\varQsa}{{q}_{\inda\indo}}
\newcommand{\varQsaaps}{{q}_{\inda'\indo'}}
\nomenclature[VQ]{$\varQsa$}{Score of action $\inda$ in observation $\indo$;}
\newcommand{\varr}{r}
\nomenclature[Vr]{$\varr$}{Reward after taking one action in the proposed algorithm;}
\newcommand{\parCF}{C^{\labF}_{\indf\indt}}
\nomenclature[PCF]{$\parCF$}{Price of electricity in feeder $\indf$ (\EUR{}/MWh);}
\newcommand{\parEEover}{\overline{E}^{\labE}_{\inde}}
\newcommand{\parEEunder}{\underline{E}^{\labE}_{\inde}}
\nomenclature[PEEover]{$\parEEover$/$\parEEunder$}{Maximum/Minimum stored energy in BSS $\inde$ (MWh);}
\newcommand{\parPD}{P^{\labD}_{\indd\inds\indt}}
\nomenclature[PPD]{$\parPD$}{Active power of demand $\indd$ (MW);}
\newcommand{\parPEeover}{\overline{P}^{\labE}_{\inde}}
\nomenclature[Ppe]{$\parPEeover$}{Maximum charging active power in BSS $\inde$ (MW);}
\newcommand{\parPR}{\tau_{\inds}}
\nomenclature[Ptau]{$\parPR$}{Probability of scenario $\inds$;}
\newcommand{\parUE}{U^{\labE}_{\inde\inds\indt}}
\nomenclature[PUE]{$\parUE$}{Availability of BSS $\indb$;}
\newcommand{\parpv}{{P}^{\labV}_{\inds\indt\indv}}
\nomenclature[Vpv]{$\parpv$}{Active power of solar unit $\indv$ (MW);}
\newcommand{\parpw}{{P}^{\labW}_{\inds\indt\indw}}
\nomenclature[Vpw]{$\parpw$}{Active power of wind unit $\indv$ (MW);}
\newcommand{\paretace}{\eta^{\labEC} _{\inde}}
\newcommand{\paretade}{\eta^{\labED}_{\inde}}
\nomenclature[Petac]{$\paretace$/$\paretade$}{Charging/Discharging efficiency of BSS $\inde$;}
\newcommand{\paralpha}{\alpha}
\nomenclature[Palpha]{$\paralpha$}{Learning rate;}
\newcommand{\pargamma}{\gamma}
\nomenclature[Pgamma]{$\pargamma$}{Discount factor;}
\newcommand{\parAGTNUM}{AGT\_NUM}
\newcommand{\parREWEPS}{REW\_EPS}
\newcommand{\parITRLIM}{ITR\_LIM}

\usepackage{hyperref}
\begin{document}
\title{A Machine Learning Approach for Prosumer Management in Intraday Electricity Markets
\thanks{\noindent This work was financially supported by the Swedish Energy Agency (Energimyndigheten) under Grant 3233. The required computation is performed by computing resources from the Swedish National Infrastructure for Computing (SNIC) at PDC center for high performance computing at KTH Royal Institute of Technology which was supported by the Swedish Research Council under Grant 2018-05973. (Corresponding author: Saeed Mohammadi.)}
}
\author{\IEEEauthorblockN{Saeed~Mohammadi\orcidicon{0000-0003-1823-9653}} 
\IEEEauthorblockA{\textit{School of Electrical Engineering and Computer Science} \\ 
\textit{KTH Royal Institute of Technology}\\
Stockholm, Sweden \\
saeedmoh@kth.se}
\and
\IEEEauthorblockN{Mohammad~Reza~Hesamzadeh\orcidicon{0000-0002-9998-9773}}
\IEEEauthorblockA{\textit{School of Electrical Engineering and Computer Science} \\ 
\textit{KTH Royal Institute of Technology}\\
Stockholm, Sweden \\
mrhesamzadeh@kth.se}
} %
\maketitle
\begin{abstract}
Prosumer operators are dealing with extensive challenges to participate in short-term electricity markets while taking uncertainties into account. Challenges such as variation in demand, solar energy, wind power, and electricity prices as well as faster response time in intraday electricity markets. Machine learning approaches could resolve these challenges due to their ability to continuous learning of complex relations and providing a real-time response. Such approaches are applicable with presence of the high performance computing and big data. To tackle these challenges, a Markov decision process is proposed and solved with a reinforcement learning algorithm with proper observations and actions employing tabular Q-learning. Trained agent converges to a policy which is similar to the global optimal solution. It increases the prosumer's profit by 13.39\% compared to the well-known stochastic optimization approach.
\end{abstract}
\vspace{12mm}
\printnomenclature
\section{Introduction}\label{sec:introduction}
\begin{figure}[!h]
\centering
\includegraphics{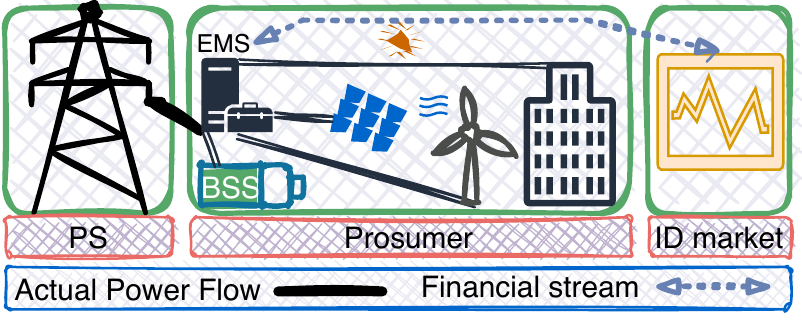}
\caption{Structure of prosumer in the proposed model. BSS: battery storage system, EMS: energy management system, ID: Intraday, PS: power system grid.}
\label{fig:structure}
\end{figure}

\IEEEPARstart{I}{ntraday} electricity markets could be categorized into continuous trading and discrete auctions. As the names imply, continuous trading is first-come-first-serve and bids and offers are matched continuously without applying any auction. This type of trades are used in Elbas market in 10 European countries now (the Nordics, Baltics, Germany, Belgium and the Netherlands) as well as Germany \cite{nordpoolelbas}. On the other hand, discrete auctions apply the same price to all market participants by employing an auction in each time interval. For instance Germany introduced both discrete auctions and continuous trading. Therefore, it is critical to decrease response time in the continuous intraday markets. On the other hand, finding an optimal solution in deregulated electricity market is challenging due to stochastic nature of this problem. Structure of the modeled prosumer, unit who both produce and consume electricity, is shown in Fig. \ref{fig:structure}. Prosumers bring different uncertainties such as variation in solar energy and wind power which adds to the existing uncertainties such as deviation in the electricity price and demand.

In recent studies, machine learning approaches are employed in electricity markets mostly to predict the uncertain parameters as in \cite{mohammadi2020review}, \cite{farooqi2019machine}, and \cite{VOYANT2017569}. Specifically, reinforcement learning (RL) approaches are studied due to their generality as discussed in \cite{mohammadi2020review}, \cite{zhang2020deep}, and \cite{RINGLER2016205}. Besides, RL is used for energy management as in \cite{mbuwir2017reinforcement}. These approaches are applicable and scalable in presence of high performance computing (HPC) and big data.
This paper focuses on using machine learning approaches to manage prosumers with battery energy storage systems (BSSs) to trade in the continuous intraday markets with the proceeding uncertainties. Using the publicly available historical data, a responsive approach is developed for continuous biding in the intraday markets to balance the prosumer's active power in real-time without defining several scenarios which is a great advantage compared to the well-known stochastic optimization approach. A Markov decision process (MDP) is demonstrated and a RL algorithm (tabular Q-learning) is employed to train a RL-based network (agent) to bid/offer in the continuous intraday markets. Performance of the proposed approach is presented in the Elbas intraday market adopting the historical data for demands, solar energy, wind power, and cost of electricity in 2018-2019 period in Stockholm, Sweden.
\subsection{contributions}
The main contributions of this paper are the following:
(i) applying tabular Q-learning for prosumer management in the intraday electricity markets,
(ii) comparing the trained agent (trained with the historical data) with the optimal solutions from stochastic and deterministic optimization, and 
(iii) proposing an MDP approach to solve the prosumer management problem with RL.
The rest of this paper is organized as follows.
Section \ref{sec:formulation} formulates the prosumer management problem and explains the proposed MDP algorithm to train the agent. 
Section \ref{sec:results} presents operation of a trained agent for a two weeks period and compares the results with the benchmark approaches, i.e. deterministic and stochastic optimization.
Section \ref{sec:conclusion} concludes the paper.

\section{Methodology}\label{sec:formulation}
As illustrated in Fig. \ref{fig:structure}, the prosumer management problem in the intraday continuous market is formulated employing the well-known stochastic optimization in \eqref{eq:opt} with set of decision variables $\setX=\{\varpf$, $\varuec$, $\varuecp$, $\varued$, $\varuedp$, $\vareet\}$. Cost of electricity, wind power, solar power, and demand for scenario $\inds$ are presented by $\parCF$, $\parpw$, $\parpv$, and $\parPD$ respectively which represent the uncertain parameters.
\begin{subequations} \label{eq:opt}
\begin{align}
&\begin{aligned}[b]
&\underset{\setX}{\text{Minimize }} 
{\textstyle \sum}_{\indf\inds\indt} \parPR  \parCF \, \varpf 
\end{aligned}  \!\! \label{eq:opt:obj} \\ 
&\text{Subject to: } {\textstyle \sum}_{\indf \in \setF_\indb}\varpf + {\textstyle \sum}_{\indw \in \setW_\indb}\parpw + {\textstyle \sum}_{\indv \in \setV_\indb}\parpv +\nonumber\\
&\;\;\;\;\;\;\;\;\;\;\;\;\;\;\;\;\,{\textstyle \sum}_{\inde \in \setE_\indb}  ( \varped \m \varpec \! ) = {\textstyle \sum}_{\indd \in \setD_\indb}  \parPD \label{eq:opt:PB_p}\\
&\;\;\;\;\;\;\;\;\;\;\;\;\;\;\;\;\, \varpec \!=\! 0.5 \varuec \parPEeover \p \varuecp \parPEeover  \!\! \label{eq:opt:pec} \\
&\;\;\;\;\;\;\;\;\;\;\;\;\;\;\;\;\, \varped \!=\! 0.5\varued\parPEeover \p \varuedp \parPEeover  \!\! \label{eq:opt:ped} \\
&\;\;\;\;\;\;\;\;\;\;\;\;\;\;\;\;\, \varuec \p \varuecp \p \varued \p \varuedp \! \le \! \parUE \!\! \label{eq:opt:uec_ued_lim} \\
&\;\;\;\;\;\;\;\;\;\;\;\;\;\;\;\;\, \vareet \m \vareetminusone \!=\! \paretace \varpec \m \varped/\paretade \label{eq:opt:ee} \\
&\;\;\;\;\;\;\;\;\;\;\;\;\;\;\;\;\, \parEEunder \! \le \! \vareet \! \le \! \parEEover \label{eq:opt:ee_lim}\\
&\;\;\;\;\;\;\;\;\;\;\;\;\;\;\;\;\, \varuec, \varuecp, \varued, \varuedp\in\{0,1\} \label{eq:opt:u}
\end{align}
\end{subequations}

The prosumer management problem \eqref{eq:opt} is a mixed integer linear programming (MILP) problem and could be solved with the available commercial solvers. Evidently, the size of the problem and solution time gradually increases by employing more scenarios for $\parCF$, $\parpw$, $\parpv$, and $\parPD$. On the other hand, the prosumer requires faster response to participate in the continuous intraday market. Therefore, a limited number of scenarios are generated to get a reasonable solution time and accordingly solve it using the available computational resources. A proper number of scenarios with probabilities $\parPR$ are required to get the best solution. This solution is used as a benchmark to validate results of the trained agent.

Objective of \eqref{eq:opt} is to minimize total operation cost of the prosumer by buying $\varpf$ MW from the intraday market in \eqref{eq:opt:obj}. Power balance equation to meet the demand $\parPD$ as written in \eqref{eq:opt:PB_p}. When demand is more than the available power, i.e. $\parPD \ge \parpv\p\parpw$, the prosumer either buys electricity from the market $\varpf\ge0$ or employs stored energy $\varped\ge0$. On the other hand, when generation is more than the demand , i.e. $\parpv\p\parpw\ge\parPD$, the prosumer either sells to the market $\varpf\le0$ or charges the BSS $\varpec\ge0$. There are five options for charging/discharging the BSS employing four binary variables in \eqref{eq:opt:pec}, \eqref{eq:opt:ped}, and \eqref{eq:opt:u}. Only one of these binary variables $\varuecp$, $\varuec$, $\varued$, and $\varuedp$ could be one as enforced by \eqref{eq:opt:u}. It indicates five charging/discharging options $(\varpec-\varped)/\parPEeover=$ $1$, $0.5$, $0$, $\m0.5$, and $\m1$. It is mandatory to have a discrete charging/discharging power to solve this problem as we are using tabular Q-learning with a discrete action space. Problem \eqref{eq:opt} should be solved considering all uncertainties which adds to the computation burden and solution time. The proposed machine learning approach solves this issue since it works in real-time once the agent is well-trained on the historical data.
\subsection{Sequential process} \label{sub:sequential}
The prosumer management problem \eqref{eq:opt} is formulated as a sequential decision making process employing the MDP and the following elements.
\subsubsection{Observations} \label{sub:observation}
Observations $\indo$ of the studied prosumer are day (0-365), hour \{(00-01),(01-02),...,(23-00)\}, stored energy (0-100\%), solar energy, wind power, price of electricity, and demand. These observations build the state space which describes the current state of the prosumer and changes based on previous actions.
\subsubsection{Action space} \label{sub:action}
Action space $\setA$ includes buy/sell from/to the intraday market and charge/discharge active power of the prosumer. Which are five actions in total $\setA =\{\m1,\m0.5,0,0.5,1\}$. The actions $\inda$ represent discrete values for $(\varpec\m\varped)/\parPEeover$. It can be written as $0.5\varuec+\varuecp-0.5\varued-\varuedp$. For instance action $\inda=0.5$ represents charging with $\varpec=0.5\parPEeover$ and similarly for other actions.
One of these five actions ($\conNA=5$) should be taken by the prosumer based on the current observation $\indo$.
\subsubsection{Cost/reward} \label{sub:reward}
Cost/reward is price of the exchanged power ($\parCF \, \varpf$) with grid in the intraday market which should be minimized in total by charging/discharging the BSS and efficient use of the generated power in the prosumer.
These costs/rewards are employed in training process by minimizing total cost at each time step.
Stored energy is limited in \eqref{eq:opt:ee_lim} and actions that violate this constraint should be avoided. 
To illustrate this, another action is used to avoid this violation and if all actions cause violating this constraint the total cost is penalized by a relatively large number.
\subsubsection{Tabular Q-learning} \label{sub:tabq}
As explained before, the agent should take the best action based on the current observation $\indo$. 
To do so, actions with highest score $\varQsa$ value are selected in MDP using a table of $\varQsa$ values.
Therefore, the agent will take action $\inda$ when $\varQsa=\max_{\inda'\in\setA'}\varQsaaps$ for observation $\indo$.
Presented MDP requires $\varQsa$ values in the current observation for all valid actions which is obtained by executing tabular Q-learning method. Proposed MDP in Algorithm \ref{algorithm} is implemented in Python to train agents with the best approximation of the $\varQsa$ table.
\begin{algorithm}[htbp]\label{algorithm}
 \SetAlgoLined
 \KwData{Historical electricity market data from Nord Pool \cite{nordpooldatabase}}
 \KwResult{Trained agents saved for further use}
 Initialize $\varQsa$ table\;
 \While{number of saved agents less than \textnormal{\parAGTNUM}}{
  Discover observation $\indo$\;
  \While{there are more actions available}{
   Select a random action $\inda$ from remaining actions\;
   Find out corresponding reward $\varr$ and next observation $\indo'$\;
   \eIf{$\indo'$ does not violate constraints}{
    break\;
   }{
    penalize the reward $\varr$\;
   }
  }
  Update $\varQsa$ with Bellman optimality equation: 
  \begin{equation}\label{eq:bellman}
   \varQsa \xleftarrow{} (1\m\paralpha)\varQsa \p \paralpha(\varr \p \pargamma \max_{\inda'\in\setA'}\varQsaaps)
  \end{equation}\\
  Initialize the best reward and saved reward\;
  \While{the best reward is more than \textnormal{\parREWEPS}}{
   Initialize the best reward\;
   \While{iteration is less than \textnormal{\parITRLIM}}{
    Let the trained agent to perform in the trained period\;
    \If{total reward is more than best reward}{
      best reward = total reward\;
    }
   }
  }
  \If{the best reward is more than saved reward}{
   save the trained agent\;
   saved reward = the best reward\;
  }
 }
  \caption{Proposed MDP approach}
\end{algorithm}

Hyper-parameters of Algorithm \ref{algorithm} have to be selected in the training process such as \parITRLIM, \parREWEPS, \parAGTNUM, $\paralpha$, and $\pargamma$ which are iteration limit, reward epsilon, maximum number of agents, learning rate, and discount factor respectively. In Bellman optimality equation \eqref{eq:bellman}, the learning rate $0\le\paralpha\le1$ is employed to define effect of new or historical $\varQsa$ values. Higher learning rate increases effect of historical data and vice versa. It is used to avoid instability in rapid changes. In addition, a discount factor $0\le\pargamma\le1$ is applied to escape from infinite loops in training. For instance, when a single action $\varr$ is available in observation $\indo$ which takes us to the observation $\indo'$ and similarly there is a single action $\varr'$ in observation $\indo'$ which takes us back to observation $\indo$ over and over. This is an example of infinite loops, which will not happen using the discount factor, we proposed above.

\section{Results and discussion}\label{sec:results}
\begin{figure}
    \centering
    \input{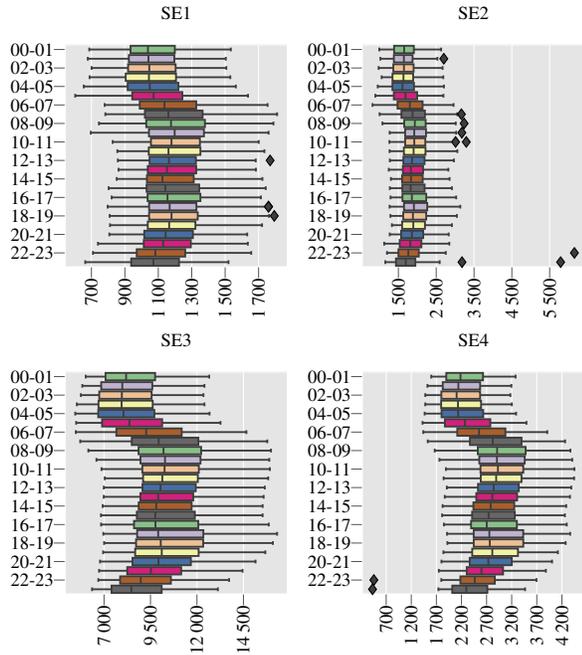}
    \caption{Power consumption in four regions of Sweden in 2019}
    \label{fig:data_consumption}
\end{figure}

The proposed Algorithm in Fig. \ref{algorithm} is applied to one prosumer with volatile demands, solar energy, wind power, and electricity prices. The second Modern-Era Retrospective analysis for Research and Applications (MERRA-2) is used for solar irradiation data which are NASA atmospheric analysis publicly available in \cite{bosilovich2015merra}. The irradiation data are used to calculated output of the solar units for 2019. The remaining data (i.e. demands, wind power, and market price) is from Nord Pool database available in \cite{nordpooldatabase} in Sweden for 2019 (2018 for wind power). The uncertain parameters of the prosumer sensitive to date, time, and place. For instance changes in demands are shown in Fig. \ref{fig:data_consumption} for four regions SE1-SE4.

\begin{figure}[htbp]
    \centering
\begin{tikzpicture}

\definecolor{color0}{rgb}{0.917647058823529,0.917647058823529,0.949019607843137}
\definecolor{color1}{rgb}{0.298039215686275,0.447058823529412,0.690196078431373}
\definecolor{color2}{rgb}{0.333333333333333,0.658823529411765,0.407843137254902}

\begin{axis}[
axis background/.style={fill=color0},
axis line style={white},
legend cell align={left},
legend style={fill opacity=0.8, draw opacity=1, text opacity=1, draw=none, fill=color0},
tick align=outside,
tick pos=left,
title={\scriptsize Convergence Rate},
x grid style={white},
xlabel={\scriptsize Number of Scenarios},
xmajorgrids,
xmin=1, xmax=40,
xtick style={color=white!15!black},
xtick={2,4,6,8,10,12,14,16,18,20,22,24,26,28,30,32,34,36,38,40},
xticklabel style = {font=\scriptsize ,rotate=90.0},
y grid style={white},
ylabel={\scriptsize OC (\EUR{})},
ymajorgrids,
ymin=0, ymax=1000,
yticklabel style = {font=\scriptsize},
ytick style={color=white!15!black},
width=\linewidth,
height=0.4\linewidth,
]
\addplot [line width=0.7pt, color1]
table {%
1	265.950518
2	265.950518
3	265.950518
4	265.950518
5	265.950518
6	265.950518
7	265.950518
8	265.950518
9	265.950518
10	265.950518
11	265.950518
12	265.950518
13	265.950518
14	265.950518
15	265.950518
16	265.950518
17	265.950518
18	265.950518
19	265.950518
20	265.950518
21	265.950518
22	265.950518
23	265.950518
24	265.950518
25	265.950518
26	265.950518
27	265.950518
28	265.950518
29	265.950518
30	265.950518
31	265.950518
32	265.950518
33	265.950518
34	265.950518
35	265.950518
36	265.950518
37	265.950518
38	265.950518
39	265.950518
40	265.950518
};
\addlegendentry{\scriptsize Trained agent cost}
\addplot [line width=0.7pt, color2, dash pattern=on 7pt off 3pt]
table {%
1	544.060993
2	232.9742052
3	513.3986139
4	786.1962805
5	840.1467391
6	937.9129009
7	999.2747293
8	923.775626
9	916.4857678
10	771.4276872
11	600.3431221
12	473.2912707
13	452.6703119
14	473.7731694
15	436.9786209
16	400.173311
17	386.6188369
18	359.1031059
19	340.4634718
20	320.5680953
21	321.7359768
22	310.0221213
23	300.9645004
24	281.1401861
25	273.9803984
26	251.3948
27	242.6387367
28	238.2912426
29	235.5531834
30	234.5309817
31	237.9832521
32	224.0310777
33	224.8382406
34	225.5369309
35	227.9744243
36	242.9092714
37	243.589652
38	247.7340275
39	249.0401662
40	249.3559447
};
\addlegendentry{\scriptsize Optimal total cost}
\end{axis}

\end{tikzpicture}
    \caption{Convergence rate of the stochastic optimization approach. OC: Operation Cost}
    \label{fig:result_convergence}
\end{figure}
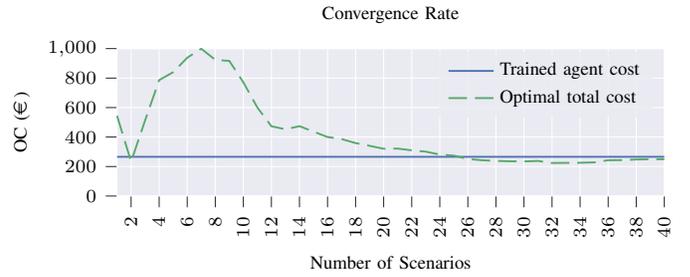

The problem \eqref{eq:opt} parameters are $\paretace=\paretade=0.9$, $\parEEunder=0\%$, $\parEEover=100\%$, and $\parPR=1/\conNS$, $\parPEeover=10$ MW.
First, optimal solution is obtained by solving \eqref{eq:opt} with CPLEX 12.8 \cite{ilog2018cplex} solver in the GAMS 25.1.3 software \cite{Bussieck2004gams} which is used as a benchmark in Section \ref{sec:results}. This problem is solved with different number of scenarios to study effect of the scenario set on operation cost. Total operation cost (${\textstyle \sum}_{\indf\inds\indt} \parPR  \parCF \, \varpf$) is shown in Fig. \ref{fig:result_convergence} to visualize convergence rate of the stochastic optimization approach by changing number of scenarios. The costs are less volatile for problems with 30 or more number of scenarios. Therefore, 30 scenarios are sufficient for stochastic optimization approach.

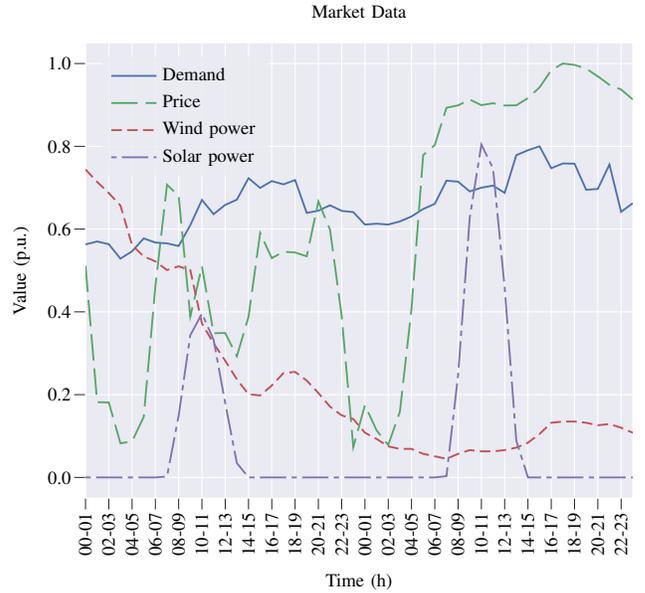
\begin{figure}[htbp]
    \centering
\begin{tikzpicture}

\definecolor{color0}{rgb}{0.917647058823529,0.917647058823529,0.949019607843137}
\definecolor{color1}{rgb}{0.298039215686275,0.447058823529412,0.690196078431373}
\definecolor{color2}{rgb}{0.333333333333333,0.658823529411765,0.407843137254902}
\definecolor{color3}{rgb}{0.768627450980392,0.305882352941176,0.32156862745098}
\definecolor{color4}{rgb}{0.505882352941176,0.447058823529412,0.698039215686274}

\begin{axis}[
axis background/.style={fill=color0},
axis line style={white},
legend cell align={left},
legend style={fill opacity=0.8, draw opacity=1, text opacity=1, at={(0.03,0.97)}, anchor=north west, draw=none, fill=color0},
tick align=outside,
tick pos=left,
title={\scriptsize Market Data},
x grid style={white},
xlabel={\scriptsize Time (h)},
xmajorgrids,
xmin=0, xmax=47,
xtick style={color=white!15!black},
xtick={0,2,4,6,8,10,12,14,16,18,20,22,24,26,28,30,32,34,36,38,40,42,44,46},
xticklabel style = {font=\scriptsize ,rotate=90.0},
xticklabels={00-01\!\!,02-03\!\!,04-05\!\!,06-07\!\!,08-09\!\!,10-11\!\!,12-13\!\!,14-15\!\!,16-17\!\!,18-19\!\!,20-21\!\!,22-23\!\!,00-01\!\!,02-03\!\!,04-05\!\!,06-07\!\!,08-09\!\!,10-11\!\!,12-13\!\!,14-15\!\!,16-17\!\!,18-19\!\!,20-21\!\!,22-23\!\!},
y grid style={white},
ylabel={\scriptsize Value (p.u.)},
ymajorgrids,
ymin=-0.05, ymax=1.05,
ytick style={color=white!15!black},
ytick={-0.2,0,0.2,0.4,0.6,0.8,1,1.2},
yticklabel style = {font=\scriptsize},
yticklabels={−0.2\!\!,0.0\!\!,0.2\!\!,0.4\!\!,0.6\!\!,0.8\!\!,1.0\!\!,1.2\!\!},
width=\linewidth,
]
\addplot [line width=0.7pt, color1]
table {%
0 0.563
1 0.57
2 0.5635
3 0.5285
4 0.5465
5 0.5775
6 0.5675
7 0.5655
8 0.559
9 0.6095
10 0.6705
11 0.636
12 0.6585
13 0.671
14 0.723
15 0.6995
16 0.716
17 0.708
18 0.7185
19 0.639
20 0.6445
21 0.6575
22 0.644
23 0.641
24 0.611
25 0.613
26 0.611
27 0.6185
28 0.6305
29 0.6485
30 0.661
31 0.717
32 0.7145
33 0.691
34 0.7
35 0.705
36 0.6875
37 0.7785
38 0.7905
39 0.8
40 0.747
41 0.7585
42 0.758
43 0.695
44 0.697
45 0.756
46 0.6415
47 0.6625
};
\addlegendentry{\scriptsize Demand}
\addplot [line width=0.7pt, color2, dash pattern=on 10pt off 3pt]
table {%
0 0.511469628845984
1 0.181874283699198
2 0.181151370889535
3 0.082359164242264
4 0.0872256016926739
5 0.146116547650533
6 0.461253636604073
7 0.707237944106497
8 0.680155161773781
9 0.388662611302125
10 0.510200123424138
11 0.348391078198008
12 0.348920038790443
13 0.292762055893503
14 0.387763378294984
15 0.590390549237415
16 0.5298950894825
17 0.54523494666314
18 0.543436480648858
19 0.534214934320727
20 0.666613770607423
21 0.59851891034118
22 0.390284757118928
23 0.0729612977166535
24 0.174521731464339
25 0.113708895353963
26 0.0782685356607599
27 0.15851185753328
28 0.407881512827294
29 0.778700520144583
30 0.803173763554615
31 0.893132328308208
32 0.898897998765759
33 0.912933086485057
34 0.899426959358194
35 0.904117076611126
36 0.898527726351054
37 0.899074318963237
38 0.916336066296394
39 0.94243145552323
40 0.983637485673984
41 1
42 0.996597020188663
43 0.987781010314732
44 0.969055805342502
45 0.94837344617826
46 0.937212377677863
47 0.913285726880014
};
\addlegendentry{\scriptsize Price}
\addplot [line width=0.7pt, color3, dash pattern=on 4pt off 2pt]
table {%
0 0.744
1 0.714
2 0.687
3 0.657
4 0.561
5 0.534
6 0.522
7 0.501
8 0.51
9 0.501
10 0.372
11 0.324
12 0.282
13 0.237
14 0.201
15 0.198
16 0.222
17 0.252
18 0.255
19 0.234
20 0.204
21 0.171
22 0.15
23 0.141
24 0.108
25 0.093
26 0.075
27 0.069
28 0.069
29 0.057
30 0.051
31 0.045
32 0.057
33 0.066
34 0.063
35 0.063
36 0.066
37 0.072
38 0.084
39 0.105
40 0.132
41 0.135
42 0.135
43 0.132
44 0.126
45 0.129
46 0.12
47 0.108
};
\addlegendentry{\scriptsize Wind power}
\addplot [line width=0.7pt, color4, dash pattern=on 2pt off 2pt on 10pt off 2pt]
table {%
0 0
1 0
2 0
3 0
4 0
5 0
6 0
7 0.00226892
8 0.14928108
9 0.343513622
10 0.395937798
11 0.33282978
12 0.17965603
13 0.034599298
14 0
15 0
16 0
17 0
18 0
19 0
20 0
21 0
22 0
23 0
24 0
25 0
26 0
27 0
28 0
29 0
30 0
31 0.00306564
32 0.247038624
33 0.627431722
34 0.804256798
35 0.746775182
36 0.458419698
37 0.08848788
38 0
39 0
40 0
41 0
42 0
43 0
44 0
45 0
46 0
47 0
};
\addlegendentry{\scriptsize Solar power}
\end{axis}

\end{tikzpicture}
    \caption{Market data for testing days}
    \label{fig:result_market_data}
\end{figure}

Market data for the first two days of January are shown in Fig. \ref{fig:result_market_data}. All values are in per unit. All active power values are divided by $1,000$ MW as base and market price is divided by its maximum value ($567.15$ SEK or $56.71$ \EUR{}).

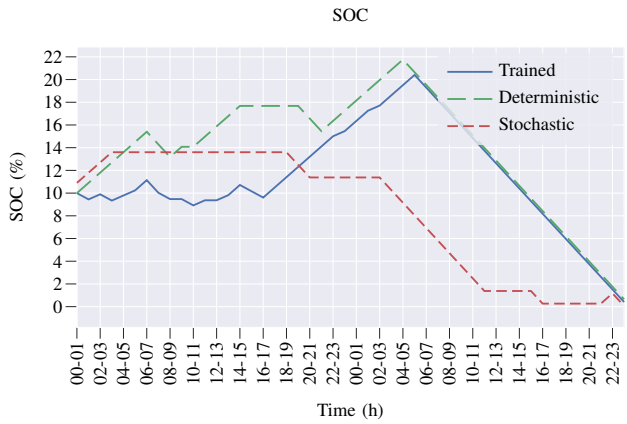
\begin{figure}[htbp]
    \centering
\begin{tikzpicture}

\definecolor{color0}{rgb}{0.917647058823529,0.917647058823529,0.949019607843137}
\definecolor{color1}{rgb}{0.298039215686275,0.447058823529412,0.690196078431373}
\definecolor{color2}{rgb}{0.333333333333333,0.658823529411765,0.407843137254902}
\definecolor{color3}{rgb}{0.768627450980392,0.305882352941176,0.32156862745098}
\definecolor{color4}{rgb}{0.505882352941176,0.447058823529412,0.698039215686274}

\begin{axis}[
axis background/.style={fill=color0},
axis line style={white},
legend cell align={left},
legend style={fill opacity=0.8, draw opacity=1, text opacity=1, draw=none, fill=color0},
tick align=outside,
tick pos=left,
title={\scriptsize SOC},
x grid style={white},
xlabel={\scriptsize Time (h)},
xmajorgrids,
xmin=0, xmax=47,
xtick style={color=white!15!black},
xtick={0,2,4,6,8,10,12,14,16,18,20,22,24,26,28,30,32,34,36,38,40,42,44,46},
xticklabel style = {font=\scriptsize ,rotate=90.0},
xticklabels={00-01\!\!,02-03\!\!,04-05\!\!,06-07\!\!,08-09\!\!,10-11\!\!,12-13\!\!,14-15\!\!,16-17\!\!,18-19\!\!,20-21\!\!,22-23\!\!,00-01\!\!,02-03\!\!,04-05\!\!,06-07\!\!,08-09\!\!,10-11\!\!,12-13\!\!,14-15\!\!,16-17\!\!,18-19\!\!,20-21\!\!,22-23\!\!},
y grid style={white},
ylabel={\scriptsize SOC (\%)},
ymajorgrids,
ymin=-1.82861111133334, ymax=22.8786111157778,
yticklabel style = {font=\scriptsize},
ytick style={color=white!15!black},
width=\linewidth,
height=0.6\linewidth,
ytick={0,2,4,6,8,10,12,14,16,18,20,22},
yticklabels={0\!\!,2\!\!,4\!\!,6\!\!,8\!\!,10\!\!,12\!\!,14\!\!,16\!\!,18\!\!,20\!\!,22\!\!},
]
\addplot [line width=0.7pt, color1]
table {%
0 10.0
1 9.444444444444445
2 9.894444444444444
3 9.338888888888889
4 9.788888888888888
5 10.238888888888887
6 11.138888888888888
7 10.027777777777777
8 9.472222222222221
9 9.472222222222221
10 8.916666666666666
11 9.366666666666665
12 9.366666666666665
13 9.816666666666665
14 10.716666666666665
15 10.16111111111111
16 9.605555555555554
17 10.505555555555555
18 11.405555555555555
19 12.305555555555555
20 13.205555555555556
21 14.105555555555556
22 15.005555555555556
23 15.455555555555556
24 16.355555555555554
25 17.255555555555553
26 17.705555555555552
27 18.60555555555555
28 19.50555555555555
29 20.405555555555548
30 19.294444444444437
31 18.183333333333326
32 17.072222222222216
33 15.961111111111105
34 14.849999999999994
35 13.738888888888884
36 12.627777777777773
37 11.516666666666662
38 10.405555555555551
39 9.29444444444444
40 8.18333333333333
41 7.072222222222219
42 5.961111111111109
43 4.849999999999998
44 3.7388888888888867
45 2.6277777777777755
46 1.5166666666666644
47 0.4055555555555532
};
\addlegendentry{\scriptsize Trained}
\addplot [line width=0.7pt, color2, dash pattern=on 7pt off 3pt]
table {%
0 10.0
1 10.9
2 11.8
3 12.7
4 13.6
5 14.5
6 15.4
7 14.28888889
8 13.17777778
9 14.07777778
10 14.07777778
11 14.97777778
12 15.87777778
13 16.77777778
14 17.67777778
15 17.67777778
16 17.67777778
17 17.67777778
18 17.67777778
19 17.67777778
20 16.56666667
21 15.45555556
22 16.35555556
23 17.25555556
24 18.15555556
25 19.05555556
26 19.95555556
27 20.85555556
28 21.75555556
29 20.64444444
30 19.53333333
31 18.42222222
32 17.31111111
33 16.2
34 15.08888889
35 13.97777778
36 12.86666667
37 11.75555556
38 10.64444444
39 9.53333333
40 8.42222222
41 7.31111111
42 6.2
43 5.08888889
44 3.97777778
45 2.86666667
46 1.75555556
47 0.64444444
};
\addlegendentry{\scriptsize Deterministic}
\addplot [line width=0.7pt, color3, dash pattern=on 4pt off 2pt]
table {%
0 10.9
1 11.8
2 12.7
3 13.6
4 13.6
5 13.6
6 13.6
7 13.6
8 13.6
9 13.6
10 13.6
11 13.6
12 13.6
13 13.6
14 13.6
15 13.6
16 13.6
17 13.6
18 13.6
19 12.48888889
20 11.37777778
21 11.37777778
22 11.37777778
23 11.37777778
24 11.37777778
25 11.37777778
26 11.37777778
27 10.26666667
28 9.15555556
29 8.04444444
30 6.93333333
31 5.82222222
32 4.71111111
33 3.6
34 2.48888889
35 1.37777778
36 1.37777778
37 1.37777778
38 1.37777778
39 1.37777778
40 0.26666667
41 0.26666667
42 0.26666667
43 0.26666667
44 0.26666667
45 0.26666667
46 1.16666667
47 0.05555556
};
\addlegendentry{\scriptsize Stochastic}
\end{axis}

\end{tikzpicture}
    \caption{SOC for the trained agent, optimal deterministic solution, and optimal stochastic solution}
    \label{fig:results_soc}
\end{figure}

Subsequently, an agent is trained employing this historical data with the proposed Algorithm \ref{algorithm} in Section \ref{sec:formulation}. The algorithm hyper-parameters are \parITRLIM$=20$, \parREWEPS$=0.4$, \parAGTNUM$=20$, $\paralpha=0.2$, and $\pargamma=0.9$. Stored energy $\vareet$ in percent as state of charge (SOC) of the trained agent for the same two days are shown in Fig. \ref{fig:results_soc} and compared to the benchmarks solutions. Our first benchmark provides deterministic solution. It is optimal solution of \eqref{eq:opt} with only one scenario ($\conNS=1, \parPR=1$) where the prosumer knows the exact values for the uncertain parameters shown in Fig. \ref{fig:result_market_data}. Our second benchmark is stochastic optimization solution. Where $30$ scenarios with the same probabilities ($\conNS=30, \parPR=1/\conNS$) are considered for the uncertain parameters as shown in Fig. \ref{fig:result_convergence}. The trained agent charges the BSS unit when the price of electricity is relatively lower and sells the extra generation to the intraday electricity market when the price is relatively higher.

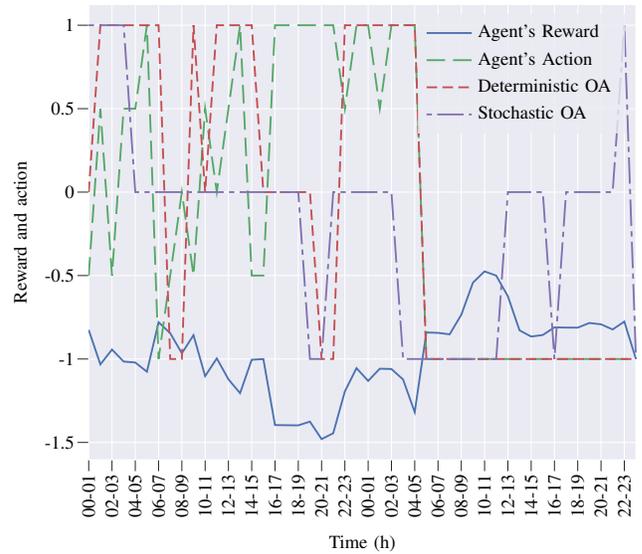
\begin{figure}[htbp]
    \centering
\begin{tikzpicture}

\definecolor{color0}{rgb}{0.917647058823529,0.917647058823529,0.949019607843137}
\definecolor{color1}{rgb}{0.298039215686275,0.447058823529412,0.690196078431373}
\definecolor{color2}{rgb}{0.333333333333333,0.658823529411765,0.407843137254902}
\definecolor{color3}{rgb}{0.768627450980392,0.305882352941176,0.32156862745098}
\definecolor{color4}{rgb}{0.505882352941176,0.447058823529412,0.698039215686274}

\begin{axis}[
axis background/.style={fill=color0},
axis line style={white},
legend cell align={left},
legend style={fill opacity=0.8, draw opacity=1, text opacity=1, draw=none, fill=color0},
tick align=outside,
tick pos=left,
x grid style={white},
xlabel={\scriptsize Time (h)},
xmajorgrids,
xmin=0, xmax=47,
xtick style={color=white!15!black},
xtick={0,2,4,6,8,10,12,14,16,18,20,22,24,26,28,30,32,34,36,38,40,42,44,46},
xticklabel style = {font=\scriptsize ,rotate=90.0},
xticklabels={00-01\!\!,02-03\!\!,04-05\!\!,06-07\!\!,08-09\!\!,10-11\!\!,12-13\!\!,14-15\!\!,16-17\!\!,18-19\!\!,20-21\!\!,22-23\!\!,00-01\!\!,02-03\!\!,04-05\!\!,06-07\!\!,08-09\!\!,10-11\!\!,12-13\!\!,14-15\!\!,16-17\!\!,18-19\!\!,20-21\!\!,22-23\!\!},
y grid style={white},
ylabel={\scriptsize Reward and action},
ymajorgrids,
ymin=-1.20826999338799, ymax=4.248012856828,
yticklabel style = {font=\scriptsize},
ytick style={color=white!15!black},
ytick={-1,0,1,2,3,4},
yticklabels={-1.5\!\!,-1\!\!,-0.5\!\!,0\!\!,0.5\!\!,1\!\!},
width=\columnwidth,
]
\addplot [line width=0.7pt, color1]
table {%
0 0.348310817244115
1 -0.0647472449969144
2 0.112947879749625
3 -0.0305964295160011
4 -0.0423480296217932
5 -0.152472617473332
6 0.440266596138588
7 0.309606790974522
8 0.0682066941902495
9 0.285672313669153
10 -0.205387285146293
11 0.0072569095128273
12 -0.243142825043287
13 -0.409691426536331
14 -0.00853079432248963
15 -0.000885585823856123
16 -0.791663263686855
17 -0.793862082341532
18 -0.795319289429604
19 -0.750571982720621
20 -0.960257136559993
21 -0.889698360222163
22 -0.387943048576214
23 -0.10944194657498
24 -0.262306162390902
25 -0.115983073261042
26 -0.120220470774927
27 -0.245614123247818
28 -0.63690698227982
29 0.318099162479062
30 0.3132377677863
31 0.295685425876047
32 0.529935089308719
33 0.915153085955991
34 1.04986223261532
35 0.998846107862364
36 0.75199555338235
37 0.343435493063211
38 0.268944635457992
39 0.287441593934585
40 0.378700431984484
41 0.3765
42 0.375717076611126
43 0.431660301507538
44 0.415724940491933
45 0.353743295424491
46 0.448456122718857
47 0
};
\addlegendentry{\scriptsize Agent's Reward}
\addplot [line width=0.7pt, color2, dash pattern=on 7pt off 3pt]
table {%
0 1
1 3
2 1
3 3
4 3
5 4
6 0
7 1
8 2
9 1
10 3
11 2
12 3
13 4
14 1
15 1
16 4
17 4
18 4
19 4
20 4
21 4
22 3
23 4
24 4
25 3
26 4
27 4
28 4
29 0
30 0
31 0
32 0
33 0
34 0
35 0
36 0
37 0
38 0
39 0
40 0
41 0
42 0
43 0
44 0
45 0
46 0
47 0
};
\addlegendentry{\scriptsize Agent's Action}
\addplot [line width=0.7pt, color3, dash pattern=on 4pt off 2pt]
table {%
0 2
1 4
2 4
3 4
4 4
5 4
6 4
7 0
8 0
9 4
10 2
11 4
12 4
13 4
14 4
15 2
16 2
17 2
18 2
19 2
20 0
21 0
22 4
23 4
24 4
25 4
26 4
27 4
28 4
29 0
30 0
31 0
32 0
33 0
34 0
35 0
36 0
37 0
38 0
39 0
40 0
41 0
42 0
43 0
44 0
45 0
46 0
47 0
};
\addlegendentry{\scriptsize Deterministic OA}
\addplot [line width=0.7pt, color4, dash pattern=on 2pt off 2pt on 10pt off 2pt]
table {%
0 4.0
1 4.0
2 4.0
3 4.0
4 2.0
5 2.0
6 2.0
7 2.0
8 2.0
9 2.0
10 2.0
11 2.0
12 2.0
13 2.0
14 2.0
15 2.0
16 2.0
17 2.0
18 2.0
19 0.0
20 0.0
21 2.0
22 2.0
23 2.0
24 2.0
25 2.0
26 2.0
27 0.0
28 0.0
29 0.0
30 0.0
31 0.0
32 0.0
33 0.0
34 0.0
35 0.0
36 2.0
37 2.0
38 2.0
39 2.0
40 0.0
41 2.0
42 2.0
43 2.0
44 2.0
45 2.0
46 4.0
47 0.0
};
\addlegendentry{\scriptsize Stochastic OA}
\end{axis}

\end{tikzpicture}
    \caption{Optimal deterministic actions, optimal stochastic action, agent's actions, and agent's rewards. OA: Optimal Action}
    \label{fig:results_reward_action}
\end{figure}

Optimal actions ($0.5\varuec+\varuecp-0.5\varued-\varuedp$) for the deterministic solution, stochastic solution, and the trained agent are shown in Fig. \eqref{fig:results_reward_action}. Also the agent's reward $\varr$ is shown in this figure. Total operation cost (${\textstyle \sum}_{\indf\inds\indt} \parPR  \parCF \, \varpf$) for these cases are $\m544$ \EUR{}, $\m234$ \EUR{}, and $\m265$ \EUR{} respectively. Which means that the prosumer is able to make a profit, hence the negative sign of the total operation cost, by selling the generated electricity from the wind and solar units. Profit of the prosumer using the trained agent is between its profit using the deterministic and stochastic solutions ($544>265>234$). The profit of the prosumer employing the trained agent is higher than the stochastic approach and lower than the deterministic approach. This is because the deterministic approach is the global optimal solution which requires a perfect knowledge about future which is not realistic in ID continuous markets. In the stochastic approach, the prosumer is more cautious and considers different scenarios for the uncertain parameters which is more realistic. But it leads to reducing the prosumer's profit in the stochastic approach. Advantage of the trained agent is to adapt the prosumer's actions based on the real-time changes in the market which increases the total profit by $13.39$ \% compared to the stochastic solution. However, there is an opportunity to improve the trained agent as the total profit is still less than the global optimal solution (i.e. deterministic solution).

\section{Conclusion}\label{sec:conclusion}
Energy storage systems can not be managed without considering the uncertainties in the current power system. This paper focused on using machine learning approaches to deal with these uncertainties and finding the best action in intraday electricity markets. A reinforcement learning network (agent) is trained based on historical changes in demands, solar irradiation, wind power, and electricity price in Stockholm, Sweden to solve the modeled Markov decision process efficiently. Preliminary results demonstrate the trained agent convergence to a policy with high scores in validation data. The trained agent managed to find a solution better than the stochastic solution  (13.39 \%) while it is less than the deterministic solution with perfect information about the uncertain parameters in future. In our planned future works, the proposed algorithm will be used to train the agent with larger database, more elements (such as hydropower units), participation in both day-ahead and intraday markets, and different observation will be employed to improve performance of the trained agent.

\ifCLASSOPTIONcaptionsoff
  \newpage
\fi
\bibliographystyle{IEEEtran}
\bibliography{IEEEabrv,bib}
\vfill
\end{document}